\documentclass[conference]{IEEEtran}
\IEEEoverridecommandlockouts
\usepackage{cite}
\usepackage{amsmath,amssymb,amsfonts}
\usepackage{algorithm} 
\usepackage{algorithmic}
\usepackage[algo2e]{algorithm2e} 
\usepackage{graphicx}
\usepackage{textcomp}
\usepackage{xcolor}
\usepackage[acronym]{glossaries}
\usepackage{arabtex}
\usepackage{utf8}

\def\BibTeX{{\rm B\kern-.05em{\sc i\kern-.025em b}\kern-.08em T\kern-.1667em\lower.7ex\hbox{E}\kern-.125emX}}
    
\makeglossaries

\newacronym{Al-Jazeera}{Al-Jazeera}{Al-Jazeera Satellite Channel}
\newacronym{AraGPT-2}{AraGPT-2}{Arabic Generative Pre-trained Transformer 2}
\newacronym{AraGPT-3}{AraGPT-3}{Arabic Generative Pre-trained Transformer 3}
\newacronym{ArSAS}{ArSAS}{Arabic Speech-Act and Sentiment}
\newacronym{PATB}{PATB}{Penn Arabic Treebank}
\newacronym{BBC}{BBC}{British Broadcasting Corporation}
\newacronym{BERT}{BERT}{Bidirectional Encoder Representations from Transformers}
\newacronym{BPE}{BPE}{byte-pair encoding}
\newacronym{mBERT}{mBERT}{multi-lingual BERT}
\newacronym{RoBERTa}{RoBERTa}{A Robustly Optimized BERT Pre-training Approach}
\newacronym{CA}{CA}{Classical Arabic}
\newacronym{CC}{CC}{Common Crawl}
\newacronym{CC-100}{CC-100}{Common Crawl 100}
\newacronym{CNN}{CNN}{Cable News Network}
\newacronym{DA}{DA}{Dialectal Arabic}
\newacronym{EAPCOUNT}{EAPCOUNT}{English-Arabic Parallel Corpus of the United Nations Texts}
\newacronym{GPT-1}{GPT-1}{Generative Pre-trained Transformer 1}
\newacronym{GPT-2}{GPT-2}{Generative Pre-trained Transformer 2}
\newacronym{GPT-3}{GPT-3}{Generative Pre-trained Transformer 3}
\newacronym{GPT}{GPT}{Generative Pre-trained Transformer}
\newacronym{GPU}{GPU}{Graphics Processing Unit}
\newacronym{HPL}{HPL}{High-Performance Linpack} 
\newacronym{IR}{IR}{Information Retrieval} 
\newacronym{KSUCCA}{KSUCCA}{King Saud University Corpus of Classical Arabic}
\newacronym{LM}{LM}{Language model}
\newacronym{LLM}{LLM}{Large Language Model}
\newacronym{LSTM}{LSTM}{Long-Short Term Memory}
\newacronym{ME}{ME}{Middle East}
\newacronym{MGB-2}{MGB-2}{Multi-Genre Broadcast-2}
\newacronym{MSA}{MSA}{Modern Standard Arabic}
\newacronym{MEGLN}{MEGLN}{MSA, Egyptian, Gulf, Levant \& North African}
\newacronym{MRC}{MRC}{Machine Reading Comprehension}
\newacronym{MT}{MT}{Machine Translation}
\newacronym{NN}{NN}{Neural Network}
\newacronym{NQ}{NQ}{Natural Questions}
\newacronym{NLP}{NLP}{Natural Language Processing}
\newacronym{OpenITI}{OpenITI}{Open Islamicate Texts Initiative}
\newacronym{OSAC}{OSAC}{Open-Source Arabic Corpora}
\newacronym{OSCAR}{OSCAR}{Open Super-large Crawled ALMAnaCH coRpus}
\newacronym{OSIAN}{OSIAN}{Open Source International Arabic News}
\newacronym{POS}{POS}{Part-of-speech}
\newacronym{PTB}{PTB}{Penn Tree Bank}
\newacronym{QA}{QA}{Question Answering} 
\newacronym{RNN}{RNN}{Recurrent Neural Network}
\newacronym{RT}{RT}{Russia Today}
\newacronym{SM}{SM}{Sentence Match}
\newacronym{SQuAD}{SQuAD}{Stanford Question Answering Dataset}
\newacronym{TPU}{TPU}{Tensor Processing Unit}
\newacronym{TFRC}{TFRC}{TensorFlow Research Cloud}
\newacronym{KSU}{KSU}{King Saud University}
\newacronym{WER}{WER}{Word Error Rate}
\newacronym{XNLI}{XNLI}{Cross-lingual Natural Language Inference}
\newacronym{HARD}{HARD}{Hotel Arabic Reviews Dataset}

\begin{document}

\title{A Large and Diverse Arabic Corpus for Large Language Models}

\author{\IEEEauthorblockN{1\textsuperscript{st} Abbas Raza Ali}
\IEEEauthorblockA{\textit{Data Science and AI Lab} \\
\textit{New York University}\\
Abu Dhabi, UAE \\
abbas.raza.ali@gmail.com}
\and
\IEEEauthorblockN{2\textsuperscript{nd} Muhammad Ajmal Siddiqui}
\IEEEauthorblockA{\textit{Smart Nations} \\
\textit{Inception Institute of AI}\\
Abu Dhabi, UAE \\
muhammad.siddiqui@inceptioniai.org}
\and
\IEEEauthorblockN{3\textsuperscript{rd} Rema Algunaibet}
\IEEEauthorblockA{\textit{Digital \& IT} \\
\textit{Saudi Aramco}\\
Al Khobar, Saudi Arabia \\
rema.algunaibet@aramco.com}
\and
\IEEEauthorblockN{4\textsuperscript{th} Hasan Raza Ali}
\IEEEauthorblockA{\textit{CTO Office} \\
\textit{Kyndryl}\\
Dubai, UAE \\
hasan.raza.ali@kyndryl.com}
}

\maketitle

\begin{abstract}


\glspl{LLM} have ushered in a major paradigm shift in \gls{NLP}, where large pre-trained \glspl{LM} have become a fundamental component of most NLP tasks. These models are intelligent enough to find relevant and meaningful representations of a language without any supervision. They are used to fine-tune typical \gls{NLP} tasks with substantially higher precision than conventional shallow learning techniques. However, training these models requires a massively large corpus that adequately represents a language. Due to the availability of enormous corpora, English \glspl{LLM} typically perform better than their counterparts.

This effort focuses on the design and development of a large Arabic corpus. The corpus comprises over 500 GB of Arabic cleaned text, intended to improve cross-domain knowledge and downstream generalization capability of \glspl{LLM}. The corpus was employed in the training of a large Arabic \gls{LLM}. In order to assess the efficacy of the \gls{LLM}, a variety of typical \gls{NLP} tasks were fine-tuned. The fine-tuned tasks exhibited a significant boost in accuracy ranging between 4.5 and 8.5\%, when compared to those downstreamed from \gls{mBERT}. To the best of our knowledge, this is currently the largest clean and diverse Arabic corpus ever assembled.

\end{abstract}


\glsresetall
\setcode{utf8}

\section{Introduction}
\glspl{LLM} have introduced a major shift in the \gls{NLP} paradigm that enables the fine-tuning of typical \gls{NLP} tasks more effectively while reducing their complexity~\cite{Radford2019, Brown2020}. The efficacy of \gls{NLP} tasks is dependent on the scale of the language model's training, which in turn depends on the availability of enormous corpora from diverse domains~\cite{kaplan2020}. In addition, the diversity of the training data is essential to improving the generalization of the large models, from where the down-stream task, such as listed in~\cite{Ali2018}, can effectively acquire knowledge given a small amount of training data~\cite{Ali2019,Ali2019a,Brown2020}.

Arabic is the 5$^{th}$ most widely spoken language in the world despite being a low-resource language. Perhaps, the most commonly used Arabic language model is Bidirectional Encoder Representations from Transformers (BERT)~\cite{Devlin2019} which is trained on a relatively small corpus, Arabic Wikipedia, compared to the \glspl{LLM} trained for other languages. Thus, it is inadequate to represent a rich and complex language like Arabic.

In order to overcome the issue, this work presents the design and development of a 500 GB high-quality Arabic corpus that may use to train the large \glspl{LM}. The corpus comprised 20 diverse data-sources, including both existing and newly developed. The \glspl{LLM} trained on this corpus greatly elevated the accuracy of a variety of down-stream tasks, which are detailed in the following sections.

\begin{table*}[!htbp]
\caption{Arabic character-set used to standardize different data-sources} \label{tab:characterset}
\centering
\small
\begin{tabular}{|p{1.2cm}p{0.9cm}l|p{1.2cm}p{0.9cm}l|p{1.2cm}p{0.9cm}l|}
\hline
{\centering \textbf{Grapheme}} & {\centering \textbf{Arpabet}} & {\centering \textbf{Description}} & {\centering \textbf{Grapheme}} & {\centering \textbf{Arpabet}} & {\centering \textbf{Description}} & {\centering \textbf{Grapheme}} & {\centering \textbf{Arpabet}} & {\centering \textbf{Description}} \\ \hline
\<ء>    	&	   E       	&	   hamza               	&	  \<د>		&		D	    	&		dal      	&	  \<م>		&		M	 	&		meem  \\ \hline
\<ا>    	&	   AE/E    	&	   alif                	&	  \<ذ>		&		DH	    	&		Thal    	&	    \<ن>		&		N	 	&		noon  \\ \hline
\<آ>    	&	   AE:     	&		alif maddah         	&	  \<ر>		&		R	    	&		ra      	&	   \<ه>		&		HH	   	&		ha       \\ \hline
\<أ>    	&	   E	    	&	   alef + hamza above  	&	  \<ز>		&		ZH	    	&		zay     	&	   \<و>		&		W	    	&		waw  \\ \hline
\<إ>    	&	   E	    	&	   alef + hamza below  	&	  \<س>		&		S	    	&		seen    	&	   \<ى>		&		AE	    	&		alif maksura   \\ \hline
\<ئ>   	&	   E	   	&	   ya+ hamza above     	&	   \<ش>		&		SH	    	&		sheen   	&	    \<ي>		&		Y	    	&		ya   \\ \hline
\<ؤ>   	&		E	   	&		waw + hamza above   	&	  \<ص>		&		SS	    	&		sad     	&	  \<َ>    	&	    AE      	&	   fathah    \\ \hline
\<إ>		&		E	   	&		alef + hamza below  	&	   \<ض>    	&		DD	 	&		dad     	&	    \<ِ>    	&	    IH      	&	   kasra   \\ \hline
\<ئ>    	&		E	   	&		ya + hamza above    	&	  \<ط>    	&		TT	 	&		ta      	&	     \<ُ>    	&	    UH      	&	   damma    \\ \hline
\<ب>		&		B	   	&		ba                  	&	 \<ظ>		&		DH2	    	&		za      	&	  \<ً>    	&	    AE N    	&	   fathathan \\ \hline
\<ة>		&		T/H	    	&		ta marbuta          	&	  \<ع> 		&		AI   	&		ayn      	&	   \<ٍ>    	&	    IH N    	&	   kasrathan   \\ \hline
\<ت>		&		T/H	   	&		ta                  	&	  \<غ>		&		GH	    	&		ghayn    	&	  \<ٌ>    	&	    UH N    	&	   dammathan \\ \hline
\<ث>		&		TH	   	&		tha                 	&	   \<ف>		&		F		&		fa      	&	  \<ّ>    	&	     -       	&	   tashdeed     \\ \hline
\<ج>		&		ZH	    	&		jeem                	&	    \<ق>   	&		Q		&		qaf     	&	   \<ْ>    	&	    -        	&	 sakun  \\ \hline
\<ح>		&		HH	    	&		ha                  	&	    \<ك>		&		KH	 	&		kaf     	&			   \multicolumn{3}{c|}{-}		\\ \hline
  \<خ>		&		KH	    	&		kha    	&	    \<ل>		&		L		&		lam	&				\multicolumn{3}{c|}{-}	\\ \hline
\end{tabular}
\end{table*}

The paradigm shift began with the introduction of \gls{BERT} in 2018 by Google~\cite{Devlin2019}. Google's \gls{BERT} outperformed the previous state-of-the-art \gls{LSTM} based \gls{RNN} models with a significant improvement of up to 7.7\% on typical \gls{NLP} tasks. In fact, the major credit goes to a powerful state-of-the-art neural architecture known as Transformer with an attention mechanism~\cite{Vaswani2017}. \gls{BERT} has another variant which is trained in 104 different languages, including Arabic, known as \gls{mBERT}. Moreover, the individual \gls{BERT} models are only available for the English and Chinese languages. The Arabic model is trained on a small corpus and only validated on the sentence contradiction task. Though, Arabic \gls{mBERT} was able to achieve state-of-the-art results in 6 text classification tasks with a few-short task-specific fine-tuning.

A year later, \cite{ElJundi2019} proposed the first Arabic Universal Language Model, referred to as hULMonA, and exhibited its use for Arabic Text classification tasks. The experimental results demonstrate that hULMonA generalizes well to diverse Arabic tasks and has obtained state-of-the-art performance on Arabic Sentiment Analysis. Consequently, AraBERT~\cite{Antoun2020} and GigaBERT~\cite{Lan2020} are two \glspl{LLM} trained on a reasonably large Arabic corpus which primarily consists of \gls{MSA} dialect. On the majority of down-stream tasks, both \glspl{LLM} produced better results compared to \gls{mBERT}.

An effort similar to this work has been made for the English language with the introduction of Pile in 2020~\cite{Pile2020}. The Pile is an 825 GB large and diverse corpus composed of 22 smaller and high-quality sources. Pile encompasses broad knowledge and reasoning abilities in a number of different disciplines, making it a robust benchmark for general text modeling ability for \glspl{LLM}. The proposed Arabic corpus effort was inspired by Pile. 

The notable contributions of this work are listed as follows:
\begin{enumerate}
    \item The design and development of over 500 GB of Arabic corpus comprising of 20 distinct new and current sources.
    \item Training an Arabic \gls{LLM} on the proposed corpus on 2.7 billion parameters \gls{GPT-3} architecture.
    \item A comprehensive evaluation of the \gls{LLM}'s performance on 5 typical supervised \gls{NLP} tasks.
\end{enumerate}

\section{Corpora}
The proposed Arabic corpus consists of 20 constituent data-sources, which are listed in Table~\ref{tab:data_sources}, and their statistics are reported in Table~\ref{tab:data_volume}. The sources belong to news, academic, social, religious, cultural, and other related domains. These sources represent a sizeable proportion of most widely spoken Arabic dialects, such as \gls{MEGLN} along with \gls{CA}.

The raw corpus is compiled from diverse sources that possess different character-sets. Hence, a common character-set was defined to standardize all the sources, which are shown in Table~\ref{tab:characterset}. The character-set included 3 short vowels namely fathah, kasra, and damma; 3 long vowels, known as alef, waw, and ya; 5 Arabic diacritical marks comprising fathatan, kasratan, dammatan, shadda, and sukun along with 36 standard consonants of Arabic. The diacritical marks are crucial to the Arabic pronunciation. The inclusion of diacritical marks in the corpus may lead to the zero-short diacritization generation which is regarded as essential for nearly all speech-processing tasks~\cite{Ali2010_1, Ali2010_2, Ali2020}. The following sections briefly summarize all the sources.

\subsection{ArabicWeb16}
ArabicWeb16 is a web crawl of a large number of Arabic web-pages with extensive coverage of both \gls{MSA} and \gls{DA}~\cite{Suwaileh2016}. The corpus consists of more than 150 million web-pages crawled during the month of January 2016 and spans a wide range of disciplines and dialects. It has several forums, including seeded question-answer sites, and numerous informational pages, such as Wikipedia, that can be used to promote questions-answering research. Table~\ref{tab:arabicweb16} depicts the distribution of dialects and domains used in this corpus.

\begin{table}[ht]
\caption{Distribution of the dialects and domains of ArabicWeb16 corpus}
\label{tab:arabicweb16}
\centering
\small
\begin{tabular}{|p{0.8cm}rp{0.85cm}|p{2.1cm}rp{0.9cm}|}
\hline
\multicolumn{3}{|c}{\textbf{Dialect} (in millions)} & \multicolumn{3}{|c|}{\textbf{Category} (in thousands)} \\ \hline
\gls{MSA}	    &	119	    &	78.86\%	&	Informational	    &	113	&	12.80\%     \\  \hline
Egyptian	    &	9.2	    &	6.10\%	&	Discussion/opinion  &	295	&	33.41\%     \\ \hline
Gulf	        &	7.6	    &	5.04\%	&	Media \& News       &	93	&	10.53\%     \\ \hline
Levantine	    &	7.2	    &	4.77\%	&	Online services	    &	36	&	4.08\%      \\ \hline
Meghrebi	    &	5.1	    &	3.38\%	&	Entertainment	    &	28	&	3.17\%      \\ \hline
Others	        &	2.8	    &	1.86\%	&	Others	            &	318	&	36.01\%     \\ \hline
\textbf{Total}	&	150.9	&	100\%   &	\textbf{Total}	    &	883	&	100\%       \\ \hline
\end{tabular}
\end{table}

\subsection{OSCAR}
The \gls{OSCAR} is a multi-lingual corpus that was extracted from \gls{CC} using language classification, filtering, and cleaning. The whole collection of \gls{CC}’s is composed of petabytes of monthly snapshots collected since 2011~\cite{Suarez2020}.

\subsection{Common Crawl (CC)}
\gls{CC} is a collection of 9 billion website crawls from the year 2012 to 2017, comprising raw web-pages, metadata, and text extractions~\cite{Buck2014}. Due to the raw nature of the corpus, \gls{CC} covers content from diverse domains at the cost of data of variable quality. Therefore, well-designed extraction and filtering routines were applied to the entire corpus, elaborated in Algorithm~\ref{alg:algo1}, resulting in higher-quality output.

\subsection{1.5 Billion Words Corpus}
The corpus is an attempt to build a contemporary linguistic corpus~\cite{Khair2016}. It contains roughly 1.5 billion words, of which approximately 3 million are unique. The corpus was collected from 5 million newspaper articles from 10 different news sources published in 8 Arab countries spanning 14 years. Table~\ref{tab:billion_word}, lists the sources of the corpus, their provenance, the time period, their country of origin, and statistics.

\begin{table}[ht]
\caption{Statistics of 1.5 billion words corpus}
\label{tab:billion_word}
\centering
\small
\begin{tabular}{|p{1.4cm}|p{0.95cm}|p{1.95cm}|r|r|}
\hline
{\textbf{Source}} & {\textbf{Country}} & {\textbf{Period}} & {\textbf{Documents}} & {\textbf{Terms}} \\ \hline
Alittihad	    & 	Emirates& 	Jan'08-Jun'14    & 	349,342	    & 	932,628     \\  \hline
Echorouk Online	& 	Algeria	& 	Feb'8-May'14	& 	139,732	    & 	543,799     \\  \hline
Alriyadh	    & 	KSA	    & 	Oct'00-Dec'13	& 	858,188	    & 	1,451,320   \\  \hline
Alyaum	        & 	KSA	    & 	Jul'00-Dec'13	& 	888,068	    & 	1,319,996   \\  \hline
Tishreen	    & 	Syria	& 	Jan'4-May'14	& 	314,597	    & 	905,169     \\  \hline
Alqabas	        & 	Kuwait	& 	Jan'06-Apr'14	& 	817,274	    & 	1,260,511   \\  \hline
Almustaqbal	    & 	Lebanon	& 	Sep'03-Apr'14	& 	446,873	    & 	982,765     \\  \hline
Almasry-alyoum	& 	Egypt	& 	Dec'15-Jan'14	& 	291,723	    & 	760,511     \\  \hline
Youm-7	        & 	Egypt	& 	Jan'8-May'13	& 	1,025,027	& 	1,020,444   \\  \hline
Saba News	    & 	Yemen	& 	Dec'9-May'14	& 	92,149	    & 	255,098     \\ \hline
\multicolumn{3}{|c|}{\textbf{Total}}	            & 	5,222,973	& 	9,432,241   \\ \hline
\end{tabular}
\end{table}

\subsection{OSIAN}
The \gls{OSIAN} corpus is a compilation of 31 international Arabic news broadcasting outlets. The filtered and cleansed data comprises of ~3.5 million articles retrieved from 6 million web-pages. These articles include more than 37 million phrases and approximately 1 billion tokens ~\cite{Zeroual2019}. This effort aimed to build a balanced corpus in which the data was drawn from a wide range of reliable and freely available sources. The crawling was carried out in March 2018.  Table~\ref{tab:osian} lists the corpus' demographic and discipline statistics of the corpus.

\begin{table}[ht]
\caption{List of OSIAN crawled web-domains}
\label{tab:osian}
\centering
\small
\begin{tabular}{|p{1.6cm}|p{4.2cm}|r|}
\hline
{\textbf{Demography}} & {\textbf{Web-domain}} & {\textbf{Articles}} \\ \hline
International 	    &	un.org, euronews.com, reuters.com, namnewsnetwork.org, sputniknews.com 	&	693,629	\\ \hline 
Middle-east 	    &	aljazeera.net, alarabiya.net 	&	366,211	\\  \hline
Algeria 	        &	djazairess.com 	    &	588,514	        \\  \hline
Australia 	        &	eltelegraph.com 	&	4,614	        \\  \hline
Canada 	            &	arabnews24.ca, halacanada.ca 	&	30,135	\\  \hline
China 	            &	arabic.cctv.com 	&	1,365	        \\  \hline
Egypt 	            &	alwatanalarabi.com 	&	85,351	        \\  \hline
France 	            &	france24.com 	    &	74,718	        \\  \hline
Iran 	            &	alalam.ir 	        &	344,011	        \\  \hline
Iraq 	            &	iraqakhbar.com 	    &	28,248	        \\  \hline
Germany 	        &	dw.com 	            &	117,261	        \\  \hline
Jordon 	            &	sarayanews.com 	    &	49,461	        \\  \hline
Morocco 	        &	www.marocpress.com 	&	188,045	        \\  \hline
Palestine 	        &	al-ayyam.ps 	    &	81,495	        \\  \hline
Qatar 	            &	raya.com 	        &	8,986	        \\  \hline
Russia 	            &	arabic.rt.com 	    &	57,238	        \\  \hline
Saudi Arabia 	    &	alwatan.com.sa 	    &	1,512	        \\  \hline
Sweden 	            &	alkompis.se 	    &	33,790	        \\  \hline
Syria 	            &	syria.news 	        &	36542	        \\  \hline
Tunisia         	&	www.turess.com 	    &	495,674	        \\  \hline
Turkey 	            &	turkey-post.net, aa.com.tr 	&	76,638	\\  \hline
UAE 	            &	emaratalyoum.com 	&	25,081	        \\  \hline
UK 	                &	bbc.com 	        &	10,686	        \\  \hline
USA 	            &	arabic.cnn.com 	    &	113,557	        \\ \hline
\multicolumn{2}{|c|}{\textbf{Total}}        &	3,512,762	    \\ \hline
\end{tabular}
\end{table}

\subsection{Wikipedia and Wiki Books}
The Arabic Wikipedia and Arabic translation of English Wikipedia are two distinct sources that were retrieved between January and October 2021 respectively. The Arabic Wikipedia is predominantly written in \gls{MSA} and Egyptian \gls{DA}~\cite{Wikipeidia2020}. The English Wikipedia is translated using English-Arabic \gls{MT}, whereas the \gls{MT} was trained on \gls{MSA} dialect corpus.

Arabic Wiki Books corpus is sharded from Arabic Wikipedia~\cite{Khooli2020}, which was cleaned and formatted. This source was merged with another Arabic books source known as Al-Shamela~\cite{Belinkov2016}. Moreover, several books were excluded from the dataset owing to bad formatting.

\subsection{MGB-2}
The \gls{MGB-2} corpus is derived from the second edition of the \gls{MGB-2} Challenge~\cite{Khurana2016}. The corpus is composed of the manually captioned recorded programs of \gls{Al-Jazeera} TV spanning over 10 years. The quality of the transcription varies substantially with multiple dialects and overlapping talkers, which is a typical scenario for political debate and talk shows. According to a rough estimation, more than 70\% of the speech is in \gls{MSA} dialect and the rest is in \gls{DA}.

\subsection{Arabic News and Ajdir Corpus}
The Arabic newspaper articles corpus was collected from \gls{BBC} Arabic, EuroNews, \gls{Al-Jazeera}, \gls{CNN} Arabic, and \gls{RT} Arabic. These news articles were collected in April 2019~\cite{Saad2018}. Table~\ref{tab:arabic_news_corpus} is reporting the statistics of the 5 different sources.

Ajdir is another multi-source news corpus comprising 113 million tokens~\cite{Abdel-Ali2005}.

\begin{table}[ht]
\caption{Statistics of Arabic News corpus} \label{tab:arabic_news_corpus}
\centering
\small
\begin{tabular}{|l|l|r|r|r|r|r|}
\hline
{\textbf{Source}} & {\textbf{Documents}} & {\textbf{Tokens}} & {\textbf{Terms}} \\ \hline
\gls{BBC}	    &	212,271 &	1,764,796	&	1,076,526	\\  \hline
\gls{RT}	    &	368,920	&	3,411,451	&	2,080,985	\\  \hline
\gls{Al-Jazeera}&	249,106	&	1,525,372	&	930,477	    \\  \hline
EuroNews	    &	46,468	&	517,227	    &	315,508	    \\  \hline
\gls{CNN}	    &	30,338	&	317,260	    &	193,529	    \\ \hline
\textbf{Total}  &	907,103	&	7,536,106	&	4,597,025	\\ \hline
\end{tabular}
\end{table}

\subsection{OSAC}
The \gls{OSAC} contains a collection of journalistic text corpus that is freely accessible~\cite{Saad2010}. The corpus is composed of web documents extracted from over 25 Arabic websites, comprising roughly 113 million tokens, that are categorized into 3 distinct sources as shown in Table~\ref{tab:osac_corpus}. The \gls{BBC} and \gls{CNN} news articles belong to 7 categories including \gls{ME}, world news, business \& economy, sports, international press, science \& technology, and art \& culture. It is assembled from numerous websites belonging to 10 categories including economics, history, education \& family, religion \& fatwas, sports, health, astronomy, law, stories, and cooking recipes.

\begin{table}[ht]
\caption{\gls{OSAC} corpus statistics}
\label{tab:osac_corpus}
\centering
\small
\begin{tabular}{|l|r|r|r|}
\hline
{\textbf{Source}} & {\textbf{Documents}} & {\textbf{Tokens}} & {\textbf{Terms}} \\ \hline
\gls{BBC}	    &	4,763	&	1,860,786	&	106,733	\\ \hline 
\gls{CNN}	    &	5,070	&	2,241,348	&	144,460	\\  \hline
\gls{OSAC}	    &	22,429	&	18,183,511	&	449,600	\\ \hline
\textbf{Total}	&	32,262	&	22,285,645	&	700,793	\\ \hline
\end{tabular}
\end{table}

\subsection{\gls{CC-100}}
It is a collection of sources comprising mono-lingual data for 100+ languages, notably data for romanized languages~\cite{Wenzek2020}. It was constructed using the URLs and paragraph indices from the \gls{CC}-Net repository of 2018 \gls{CC} snapshots. Each file comprised of documents separated by double-newlines and paragraphs separated by a single newline.


\subsection{Tashkeela}
Tashkeela is a fully-diacritized Arabic corpus that is commonly used to train \gls{NLP} tasks such as automatic diacritization and disambiguation resolution~\cite{Zerrouki2017}. The corpus contains 75.6 million fully-vocalized words mainly collected from 97 books written in the \gls{MSA} and \gls{CA} dialects. The sources of the corpus along with their statistics are listed in Table~\ref{tab:tashkeela_corpus}. Additionally, a number of modest, fully-diacritized sources pertaining \gls{MEGLN} dialects were collected~\cite{Zadeh2007}.

\begin{table}[ht]
\caption{Tashkeela corpus statistics}  \label{tab:tashkeela_corpus}
\centering
\small
\begin{tabular}{|p{4.45cm}|l|r|}
\hline
{\textbf{Source}} & {\textbf{Dialect}} & {\textbf{Tokens}} \\ \hline
Shamela Library - 97 Books	        &	\gls{CA}	                        & 74,762,008	\\  \hline
Modern Books - 20	                &	\gls{MSA}	                        & 398,911     \\ \hline
Crawl from learning.aljazeera.net, al-kalema.org, enfal.de	& \gls{MSA}	    & 461,283	 	\\  \hline
Manually diacritized	            &	\gls{MSA}	                        & 7,701	    \\ \hline
\multicolumn{2}{|c|}{\textbf{Total}}			                                & 75,629,903	\\ \hline
\end{tabular}
\end{table}

\subsection{KSUCCA} 
\gls{KSUCCA} is a pioneering 50 million tokens annotated corpus representing \gls{CA} dialect from the pre-Islamic era to the $4^{th}$ Hijri~\cite{Alrabiah2013}. This corpus aims at studying the distributional lexical semantics of \emph{The Holy Quran} words. However, it can also be used for other \gls{NLP} tasks, such as \gls{IR}, \gls{MT}, and \gls{QA}. 


\subsection{EAPCOUNT and AMARA}
\gls{EAPCOUNT} is a well-known English-Arabic parallel corpus comprising 5,392,491 words compiled from two sources, including the English original text and its corresponding Arabic translation. The content of the corpus primarily includes resolutions and annual reports issued by different United Nations organizations, along with excerpts from the publications of other international institutions~\cite{Ziemski2016}. AMARA is another source of online educational video subtitles that provides multi-lingual alignment of 20 languages~\cite{Abdelali2014}.

\subsection{ArSAS and Arabic Tweets}
\gls{ArSAS} is an annotated Arabic tweets corpus generally used for sentiment analysis task~\cite{Elmadany2018}. A vast collection of 21,000 Arabic tweets were collected from a variety of domains. Moreover, the same set of tweets was categorized into 4 classes including positive, negative, neutral, and mixed. This corpus was coupled with another large corpus of sentiment analysis having 4,418,128 tweets known as Arabic Tweets Corpus~\cite{Soliman2017}.

\subsection{OpenITI}
\gls{OpenITI} is a large-scale, historical corpus of Arabic containing 1 billion words from various epoch periods. The corpus was cleaned and processed with a morphological analyzer. This corpus was combined with the second version of Al-Shamela which contained 6,111 historical Islamic books, accessible until the beginning of the month of Rajab 1433 AH.

\begin{table*}[!htbp]
\caption{Overview of the data-sources used to develop the proposed corpus}
\label{tab:data_sources}
\centering
\small
\begin{tabular}{|p{4.4cm}|l|l|l|c|c|}
\hline
{\centering \textbf{Corpus Name}} & {\centering \textbf{Language}} & {\centering \textbf{Dialect}} & {\centering \textbf{Domain}} & {\centering \textbf{Duration}} & {\centering \textbf{Reference}} \\ \hline
ArabicWeb16	                        &	Arabic \& English	&	Multi-dialect	        &	Cross	    &	Jan 2016    &	\cite{Suwaileh2016}	\\  \hline
\gls{OSCAR}	                        &	Arabic	            &	\gls{MSA} \& Egyptian	&	Cross	    &	2011       	&	\cite{Suarez2020}	\\  \hline
\gls{CC}	                        &	Arabic	            &	\gls{MEGLN}	            &	Cross	    &	2012-2017   &	\cite{Buck2014}	\\  \hline
1.5 Billion Words Corpus	        &	Arabic	            &	\gls{MEGLN}	            &	News	    &	2000-2014	&	\cite{Khair2016}	\\  \hline
\gls{OSIAN} Corpus	                &	Arabic	            &	Multi-dialect	        &	News	    &	Mar 2018	        &	\cite{Zeroual2019}	\\ \hline 
Arabic Wikipedia	                &	Arabic	            &	\gls{MSA} \& Egyptian	&	Cross	    &	Feb 2021	&	\cite{Wikipeidia2021}	\\  \hline
English Wikipedia (translated)	    &	Arabic \& English	&	\gls{MSA}   	        &	Cross	    &	Oct 2020	&	\cite{Wikipeidia2020}	\\  \hline
Arabic Wiki Books	                &	Arabic	            &	\gls{MSA}	            &	Education	&	Feb 2020	&	\cite{Khooli2020}	\\  \hline
\gls{MGB-2}	                        &	Arabic	            &	\gls{MEGLN}	            &	News	    &	2005-2015   &	\cite{Khurana2016}	\\  \hline
Arabic News and Ajdir Corpus	    &	Arabic	            &	\gls{MSA}	            &	News	    &	Apr 2019    &	\cite{Saad2018}	\\  \hline
\gls{OSAC}	                        &	Arabic	            &	Multi-dialect	        &	News	    &	2010	    &	\cite{Saad2010}	\\  \hline
\gls{CC-100}	                    &	Arabic	            &	\gls{MSA}	            &	Cross	    &	Jan-Dec 2018&	\cite{Wenzek2020}	\\  \hline
Tashkeela	                        &	Arabic	            &	\gls{MEGLN}	\& \gls{CA} &	Education	&	-	        &	\cite{Zadeh2007, Zerrouki2017}	\\ \hline 
\gls{KSUCCA}	                    &	Arabic	            &	\gls{CA}	            &	Education	&	-	        &	\cite{Alrabiah2013}	\\  \hline
\gls{EAPCOUNT} and Amara	        &	Arabic \& English	&	\gls{MSA}	            &	Cross	    &	-	        &	\cite{Ziemski2016, Alotaibi2017}	\\  \hline
\gls{ArSAS}	and Arabic Tweets Corpus&	Arabic	            &	\gls{MSA}	            &	Social	    &	-	        &	\cite{Elmadany2018}	\\  \hline
\gls{OpenITI}	                    &	Arabic	            &	\gls{CA}	            &	Religion	&	-	        &	\cite{Belinkov2016, Belinkov2019}	\\  \hline
Arabic Books Corpus	                &	Arabic	            &	Multi-dialect	        &	Education	&	-	        &	\cite{Bellah2014}	\\  \hline
Hadith Books	                    &	Arabic	            &	\gls{CA}	            &	Religion	&	9$^{th}$ century        &	\cite{Kamr2020}	\\ \hline 
Arabic Poetry	                    &	Arabic	            &	Multi-dialect	        &	Culture	    &	till 2019   &	\cite{Alhazmi2018} \cite{El-Haj2020}	\\ \hline
\end{tabular}
\end{table*}

\subsection{Arabic Books Corpus}
It is a large Arabic corpus developed as part of a research project namely \emph{A New Approach of Semi-Indexing of Text Documents}~\cite{Bellah2014}. The corpus comprises of more than 460 Arabic books. It can be used for the development of language engineering applications, \gls{IR}, and information extraction. The size of the corpus is 137 MB which contains 23,264,785 tokens.

\subsection{Hadith Books}
It contains Hadith from the 9 sahih books scraped from \emph{islambook} website. There is a total of 62,169 Hadith where Sahih Bukhari has 7,008, Sahih Muslim has 5,362, Sunan Al-Tirmidhi has 3,891, Sunan al-Nasai has 5,662, Sunan Abu-Dawud has 4,590, Sunan Ibn Maja has 4,332, Musnad Ahmad ibn Hanbal has 26,363, Maliks Muwatta has 1,594 and Sunan al Darami has 3,367 Hadiths.

\subsection{Arabic Poetry}
The Arabic Poetry corpus is the oldest and the most prominent form of Ancient Arabic poetry that is accessible today~\cite{Alhazmi2018}. The corpus has over 58,000 poems composed in the 6$^{th}$ century. This was combined with a song lyrics corpus, known as \emph{Habibi}, which is the first Arabic song lyrics corpus collected by~\cite{El-Haj2020}. This source comprises more than 30,000 Arabic song lyrics in 6 Arabic dialects by singers from 18 Arab countries. The lyrics were segmented into more than 500,000 verses having 3.5 million words.

\begin{table}[ht]
\caption{The statistics of the raw and clean sources where the text size is in GB}
\label{tab:data_volume}
\centering
\small
\begin{tabular}{|p{2.8cm}|r|r|r|}
\hline
{\centering \textbf{Corpus Name}} & {\centering \textbf{Documents}} & {\centering \textbf{Raw Text}} & {\centering \textbf{Clean Text}} \\ \hline
ArabicWeb16	                    &	3,005	&	12,000.00   &	99.80   \\ \hline
\gls{OSCAR}                     &	46	    &	80.07	    &	76.07   \\ \hline
\gls{CC}	                    &	25	    &	1,200.00	&	250.00  \\ \hline
1.5 Billion Words Corpus	    &	10	    &	16.50	    &	15.00	\\ \hline
\gls{OSIAN} Corpus	            &	24	    &	4.70	    &   4.50	\\ \hline
Arabic Wikipedia	            &	12	    &	21.20	    &	6.60	\\ \hline
English Wikipedia (translated)	&	112	    &	22.00	    &	13.00	\\ \hline
Arabic Wiki Books	            &	512	    &	5.70	    &	5.50	\\ \hline
\gls{MGB-2}	                    &	1	    &	1.30	    &	1.20	\\ \hline
Arabic News and Ajdir Corpus    &	203	    &	1.816	    &   1.80	\\ \hline
\gls{OSAC}	                    &	32,262	&	0.29	    &	0.27	\\ \hline
\gls{CC-100}	                &	1	    &	28.00	    &   26.00	\\ \hline
Tashkeela	                    &	397	    &	1.493	    &   1.34	\\ \hline
\gls{KSUCCA}	                &	410	    &	0.44	    &   0.44	\\ \hline
\gls{EAPCOUNT} and Amara        &	17,455	&	9.55	    &   9.09	\\ \hline
\gls{ArSAS} and Arabic Tweets Corpus   &	3	    &	0.770	    &   0.76	\\ \hline 
\gls{OpenITI}	                &	7,145	&	13.85	    &   13.00	\\ \hline
Arabic Books Corpus	            &	462	    &	0.234	    &   0.22	\\ \hline
Hadith Books	                &	18	    &	0.10	    &   0.90	\\ \hline
Arabic Poetry	                &	30,073	&	0.209	    &   0.09	\\ \hline
\textbf{Total}	                &	92,218  &	13,000	    &	500.24	\\ \hline
\end{tabular}
\end{table}

\section{Data Cleansing}
The corpus was collected from multiple sources spanning from high-quality to the very noisy corpus. Each source came up with its own character-set, diacritical marks, and other normalization ones. These challenges necessitated the cleansing and normalization of distinct sources possess different rules, into a standardized corpus. The text of the data-sources was standardized according to the character-sets provided in Table~\ref{tab:characterset}. However, the characters which are not listed in the character-set were regarded to be \emph{noisy character}. The \emph{hamza above} diacritical mark was normalized by combining it with its corresponding long vowel instead of separately dealing hamza as diacritical mark. Furthermore, both English and Arabic punctuation marks were found in the raw text which were normalized to Arabic punctuation marks. The punctuation marks other than ``! ? .'' were removed from the text. For instance, in Arabic, hyphen is used as a sentence marker and also Arabic version of the comma, semi-colon, and question mark was used instead of their English versions. Conversely, the Arabic digits were normalized to their corresponding English digits, since the English digits are frequently occurring in the Arabic text.

In order to further reduce lexical sparsity, all the sources were cleaned by masking the web-links, email addresses, phone numbers, and similar kinds of identification-related information with their corresponding tags. Sometimes in Arabic, the characters are repeated more than twice in a word either to emphasize or make it the title of the document, known as \emph{tatweel}. Tatweel was also stripped during the cleansing process. The tokenization was done using Farasa word segmentor~\cite{darwish2016} and the emojis were preserved in the text. The Arabic word segmentation was applied before \gls{BPE} \cite{BPE} tokenization which was found to be effective for the down-stream tasks. The data cleansing pseudocode is outlined in Algorithm~\ref{alg:algo1}.

\IncMargin{1em}
\begin{algorithm}
\small
\caption{Corpus cleansing and standardization routine} \label{alg:algo1}
    {$sentence$.replace(URLs, [link])};  \tcp*[h]{replace URLs with tag} \\
    {$sentence$.replace(Emails, [mail])};  \tcp*[h]{replace emails with its correspoding tag} \\ 
    {$sentence$.replace(html\_markup, `')};  \tcp*[h]{remove html markup} \\
    
    {$sentence$.replace(\<ـ>, `')};  \tcp*[h]{remove tatweel} \\
    
    {$sentence$ = normalize hamza above \<أؤئ>}; \\
    
    {$sentence$ = remove parentheses \& content between them}; \\
    
    {$sentence$ = standardize Arabic characters}; \\
    
    {$sentence$.replace(punctuation, `')};  \tcp*[h]{remove punctuation} \\
    
    {$sentence$.replace(noisy\_characters, `')};  \tcp*[h]{remove noisy characters} \\
    
\end{algorithm}

\section{Language Modeling}
The cleaned corpus was used to train a large Arabic \gls{LM}. The model was trained without any supervision as the next word prediction problem. In order to train the model, a \gls{GPT-3} architecture containing 2.7 billion parameters were used. The model consisted of 32 transformer decoder layers, each of which comprised of self-attention and feed-forward sub-layers, containing 32 attention heads per layer having 80 dimensions each. Moreover, a large batch size of 1 million instances was used in conjunction with a learning rate of 1.6x10$^{-4}$.

A mesh-Tensorflow implementation was used to train the model due to the unavailability of \gls{GPT-3} implementation~\cite{EleutherAI}. \cite{GPT3_Encoder} implementation was used for \gls{BPE} which converted the text into a series of integers. The input was transformed into \gls{BPE} before feeding into the model. The model was trained on 128 x \gls{TPU} v3 for 320 hours.

\section{Evaluation}
The \gls{LLM} was evaluated by fine-tuning a number of standard \gls{NLP} tasks, such as \gls{MRC}, Text Summarization, Diacritics Restoration, Sentiment Analysis, and \gls{POS} Tagger. Despite the fact that the proposed \gls{LLM} is a \gls{GPT} architecture, it was compared to the most popular LM, mBERT. Perhaps, both of them are based on the transformer architecture, however, they are fundamentally different where \gls{BERT} consists of only encoder blocks, conversely, GPT has decoder blocks.

The \gls{MRC} task was fine-tuned on the Arabic-translated version of \gls{SQuAD} comprising 48,344 question-answer pairs~\cite{Mozannar2019}. The evaluation was performed using the \gls{SQuAD} metric, which measures the proportion of correct predictions that fall within the same sentence as the ground truth answer.

Text Summarization was trained on a multi-purpose Arabic corpus of 20,291 articles with respective extractive summaries. The dataset belongs to 6 broad categories including culture, economy, local news, international news, religion, and sports. The quality of summarization was evaluated and compared with the down-stream task from \gls{mBERT} using Rough measures~\cite{Lin2004}.

The diacritics restoration task was fine-tuned on the transcript of the widely used Arabic speech recognition corpus, known as \gls{KSU}~\cite{Alsulaiman2013}. The corpus contains 10,000 rich and balanced sentences that were manually diacritized on the MSA dialect. In order to evaluate this task, \gls{WER} was computed at the word-level. 

The sentiment analysis task was down-streamed on the HARD dataset comprising of 93,700 hotel reviews~\cite{Elnagar2018}. The dataset covers both \gls{MSA} and \gls{DA}. The reviews were categorized into positive, negative, and neutral reviews. To determine the effectiveness of the fine-tuning, the accuracy metric was adopted.

\begin{table*}[ht]
\caption{List of tasks data-sources and their evaluation on the proposed \gls{LLM} vs \gls{mBERT}}
\label{tab:evaluations}
\centering
\small
\begin{tabular}{|l|l|l|c|c|}
\hline
{\centering \textbf{Task}} & {\centering \textbf{Corpus Name}} & {\centering \textbf{Size}} & {\centering \textbf{Proposed \gls{LLM}}} & {\centering \textbf{\gls{mBERT}}} \\ \hline
\gls{MRC}               & \gls{SQuAD} (translated)      & 48,344 pairs      & 94.04 & 90.23 \\ \hline
Text Summarization      & KALIMAT                       & 20,291 articles   & 41.99 & 32.08 \\ \hline
Diacritics Restoration  & \gls{KSU} transcript          & 10,000 sentences  & 98.21 & 95.39 \\  \hline
Sentiment Analysis      & \gls{HARD}                    & 24,028 comments   & 96.35 & 95.73  \\ \hline 
\gls{POS} Tagger        & Arabic \gls{PTB}              & 738,845 tokens    & 15.52 & 23.57 \\ \hline
\end{tabular}
\end{table*}

\gls{PATB} is the most commonly used dataset for \gls{POS} Tagging~\cite{Maamouri2004}. It contains approximately 300,000 pairs of Arabic words which are grouped into 33 \gls{POS} tags. The performance of the POS tagger was tested using the \gls{WER} measure.

The \gls{LLM} trained on the proposed corpus has significantly outperformed the multi-lingual \gls{mBERT} on all the down-stream \gls{NLP} tasks. The statistics of the few-shot learning on the proposed \gls{LLM} along with their comparison with the tasks fine-tuned on \gls{mBERT} are reported in Table~\ref{tab:evaluations}.

\section{Conclusion}
This paper details the design and development of the largest clean and most diverse Arabic Corpus, specifically for training large-scale \glspl{LM}. Although Arabic is the 5$^{th}$ most widely spoken language, the Arabic \glspl{LLM} are trained on a relatively small corpus which is an insufficient representation of such a rich language. Therefore, the Arabic \gls{NLP} tasks persistently struggled to perform better than English tasks, emphasizing the need for a diversified Arabic corpus sufficient for training a large Arabic \gls{LM}.

The proposed corpus is inspired by Pile. It consists of over 500 GB which intends to improve cross-domain knowledge and down-stream generalization capability of Arabic \glspl{LLM}. Furthermore, the corpus was utilized to train a \gls{GPT-3} model having 2.7 billion parameters. This model was evaluated by fine-tuning several supervised learning \gls{NLP} tasks. The evaluation proved the efficacy of the diverse corpus and \gls{LLM} trained on it. The tasks that were fined-tuned using the proposed \gls{LLM} significantly outperformed the ones down-stream on \gls{mBERT}.

\section*{Acknowledgements}
The author wishes to thank \gls{TFRC} for providing free access to Cloud \glspl{TPU}, this wouldn't have been accomplished without this program. Also a special thanks to Dr. Marcin Budka, Professor of Data Science at Bournemouth University, UK for participating in the subject discussion and review of this work.

\bibliographystyle{plain}
\bibliography{references}

\end{document}